\documentclass[twoside]{article}

\usepackage[accepted]{aistats2021}
\usepackage[round]{natbib}

\usepackage{printlen}

\usepackage[utf8]{inputenc} 
\usepackage[T1]{fontenc}    
\usepackage{hyperref}       
\usepackage{url}            
\usepackage{booktabs}       
\usepackage{amsfonts}       
\usepackage{nicefrac}       
\usepackage{microtype}      
\usepackage{xcolor}
\usepackage{calc}

\usepackage{dsfont}    
\usepackage{todonotes}
\usepackage{amsmath,amssymb}
\usepackage{amsthm}
\usepackage{multirow}
\usepackage{graphicx}
\usepackage{listings}
\usepackage{microtype}
\usepackage{graphicx}
\usepackage{wrapfig}
\usepackage{booktabs} 
\usepackage{siunitx}
\usepackage{hyperref}
\hypersetup{
    colorlinks=true,
    linkcolor=black,
    filecolor=black,      
    urlcolor=blue,
    citecolor=black
}

\usepackage{subcaption}
\usepackage{tikz}
\usetikzlibrary{positioning,fit,bayesnet}
\tikzset{
latent/.style={circle, minimum size = 8mm, thick, draw =black!100, node distance = 10mm},
observed/.style={circle, minimum size = 8mm, thick, draw =black!100, fill=black!20, node distance = 10mm},
label/.style={circle, minimum size = 8mm, thick, draw =black!100, fill=black!20, node distance = 5mm},
dag/.style={-latex, thick},
nn/.style={-latex', dashed}
}

\newcommand{\E}{\mathbb{E}}
\newcommand{\N}{\mathcal{N}\!}

\newcommand{\mean}[2]{\E_{#2} \! \left[ #1 \right]}

\renewcommand{\d}[1]{\,\text{d} #1}
\newcommand{\dt}{\d{t}}

\newcommand{\vt}[1]{\boldsymbol{#1}}

\definecolor{mygreen}{rgb}{0,0.6,0}
\definecolor{mygray}{rgb}{0.5,0.5,0.5}
\definecolor{mymauve}{rgb}{0.58,0,0.82}

\DeclareFixedFont{\ttb}{T1}{txtt}{bx}{n}{8} 
\DeclareFixedFont{\ttm}{T1}{txtt}{m}{n}{8}
\DeclareFixedFont{\ttmsmall}{T1}{txtt}{m}{n}{7}
\DeclareFixedFont{\ttbsmall}{T1}{txtt}{bx}{n}{7}
\newcommand\pythonstyle{\lstset{
language=Python,
basicstyle=\ttm,
commentstyle=\ttm,
otherkeywords={self, yield},             
keywordstyle=\ttb\color{blue},
emph={MyClass,__init__},          
emphstyle=\ttb\color{red},    
stringstyle=\color{mygreen},
commentstyle=\color{mygreen},
showstringspaces=false,            %
breaklines=true,
frame=tb,
deletekeywords={apply},
literate={-}{-}1
}}

\newcommand\smallpythonstyle{\lstset{
language=Python,
basicstyle=\linespread{0.8}\selectfont\ttmsmall,
commentstyle=\ttmsmall,
otherkeywords={self, yield},             
keywordstyle=\ttbsmall\color{blue},
emph={MyClass,__init__},          
emphstyle=\ttbsmall\color{red},    
stringstyle=\color{mygreen},
commentstyle=\color{mygreen},
showstringspaces=false,            %
breaklines=true,
frame=tb,
deletekeywords={apply}
literate={-}{-}1
}}

\lstnewenvironment{python}[1][]
{
\pythonstyle
\lstset{#1}
}
{}

\lstnewenvironment{smallpython}[1][]
{
\smallpythonstyle
\lstset{#1}
}
{}

\newcommand\pythoninline[1]{{\pythonstyle\lstinline!#1!}}
\renewcommand{\v}[1]{\pythoninline{#1}}

\begin{document}

\runningauthor{Ambrogioni, Lin, Fertig, Vikram, Hinne, Moore and van Gerven}

\setlength{\abovedisplayshortskip}{0pt plus 1pt} 
\setlength{\belowdisplayshortskip}{3pt plus 1pt minus 1pt} 
\setlength{\abovedisplayskip}{3pt plus 1pt minus 1pt} 
\setlength{\belowdisplayskip}{3pt plus 1pt minus 1pt} 
\frenchspacing

\twocolumn[

\aistatstitle{Automatic structured variational inference}

\aistatsauthor{Luca Ambrogioni$^1$ \And Kate Lin$^2$ \And Emily Fertig$^2$ \And Sharad Vikram$^2$} 

\aistatsauthor{Max Hinne$^1$ \And Dave Moore$^2$ \And Marcel van Gerven$^1$}

\aistatsaddress{$^1$Department of Artificial Intelligence,
  Donders Institute for Brain, Cognition and Behaviour \\
  Radboud University, Nijmegen, the Netherlands \\
  $^2$Google Research, San Francisco, CA, USA}
]

\begin{abstract}
Stochastic variational inference offers an attractive option as a default method for differentiable probabilistic programming. However, the performance of the variational approach depends on the choice of an appropriate variational family. Here, we introduce automatic structured variational inference (ASVI), a fully automated method for constructing structured variational families, inspired by the closed-form update in conjugate Bayesian models. These convex-update families incorporate the forward pass of the input probabilistic program and can therefore capture complex statistical dependencies. Convex-update families have the same space and time complexity as the input probabilistic program and are therefore tractable for a very large family of models including both continuous and discrete variables.  We validate our automatic variational method on a wide range of low- and high-dimensional inference problems. We find that ASVI provides a clear improvement in performance when compared with other popular approaches such as the mean-field approach and inverse autoregressive flows. We provide an open source implementation of ASVI in TensorFlow Probability.
\end{abstract}



\section{Introduction} 
The aim of probabilistic programming is to automate every aspect of probabilistic inference in arbitrary probabilistic models (programs), so that the user can focus her attention on modeling. Stochastic gradient-based variational methods have recently emerged as a powerful alternative to MCMC for inference in differentiable probabilistic programming languages~\citep{wingate2013automated, kucukelbir2017automatic, tran2016edward, kucukelbir2015automatic, bingham2019pyro}. This trend is a consequence of the increasing automatization of variational inference (VI) techniques, which evolved from requiring highly mathematically sophisticated and model-specific tools, to generic algorithms that can be applied to a broad class of problems without model-specific derivations~\citep{hoffman2013stochastic, kingma2013auto, hernandez2015probabilistic, ranganath2014black}. However, applications of VI usually require the user to specify a parameterized variational family. In general, it is relatively easy to automate the construction of the variational distribution under a mean-field approximation, where the approximate posterior distribution factorizes as a product of univariate distributions. However, while there is a substantial body of model-specific research on structured variational families, few existing methods can be used for automatically constructing an appropriate scalable structured variational approximation for an arbitrarily chosen probabilistic model. Instead, existing methods either ignore most of the prior structure of the model (e.g. ADVI with the multivariate Gaussian distribution~\citep{kucukelbir2017automatic}) or require strict assumptions such as local conjugacy (e.g. structured stochastic VI~\citep{hoffman2015stochastic}). In addition, several of these methods require the use of \emph{ad-hoc} gradient estimators or variational lower bounds~\citep{tran2015copula, ranganath2016hierarchical}. 

In this paper we introduce an automatic procedure for constructing variational approximations that incorporate the structure of the probabilistic model, while being flexible enough to capture the distribution of the observed data. The construction of these variational approximations is fully automatic and the resulting variational distribution has the same time and space complexity as the input probabilistic program. The new family of variational models, which we call the \emph{convex-update variational family}, interpolates the evidence coming from the observed data with the feedforward structure of the model prior. Specifically, the parameters of the posterior distribution of each latent variable are a convex combination of the parameters induced by the parents of that variable, and a term reflecting the influence of the data. This mimics the evidence update in the expectation parameters of conjugate exponential family models, where this posterior is given in closed form. The convex-update variational family can be trained using standard inference techniques and gradient estimators and can therefore be used as a drop-in replacement of the mean-field approach in automatic differentiation stochastic VI. We refer to this new form of fully automatic inference as \emph{automatic structured variational inference} (ASVI). A full implementation is open-sourced in TensorFlow Probability as \href{https://www.tensorflow.org/probability/api_docs/python/tfp/experimental/vi/build_asvi_surrogate_posterior}{\protect\v{tfp.experimental.build_asvi_surrogate_posterior}}.

\section{Preliminaries}
VI is used to approximate the posterior over the latent variables of a probabilistic program $p(\vt{x})$ with a member of a parameterized family of probability distributions $q(\vt{x}; \vt{\psi})$. The vector of variational parameters $\vt{\psi}$ is obtained by maximizing the evidence lower bound (ELBO):
\begin{equation} \label{eqn:elbo}
    \mathcal{L}\left[\vt{\psi} \right] = - \mean{\log{\frac{q(\vt{x}; \vt{\psi})}{L\left(\vt{y}\mid \vt{x}\right) p(\vt{x})}}}{q(\vt{x}; \vt{\psi})}~,
\end{equation}
where $L\left(\vt{y}\mid \vt{x}\right)$ is a likelihood function. The resulting variational posterior (i.e. the maximum of this optimization problem) depends on the choice of the parameterized family $q(\vt{x}; \vt{\psi})$ and is equal to the exact posterior only when the latter is included in the family. In this paper we restrict our attention to probabilistic programs that are specified in terms of conditional probabilities and densities chained together by deterministic functions:
\begin{equation}
    p(\vt{x}) = \prod_j \rho_j \left(x_j \mid \vt{\theta}_j(\vt{\pi}_j) \right) \label{eq: probabilistic program}~,
\end{equation}
where $\rho_j\left(\cdot \mid \cdot \right)$ is a family of probability distributions and
$\vt{\pi}_j \subseteq \{x_i\}_{i \neq j}$
is a subset of parent variables such that the resulting graphical model is a directed acyclic graph (DAG). The vector-valued functions $\vt{\theta}_j(\vt{\pi}_j)$ specify the value of the parameters of the distribution of the latent variable $x_j$ given the values of all its parents. 


\subsection{Convex updates in conjugate models}
Exponential family distributions play a central role in Bayesian statistics as the only families that support conjugate priors, where the posterior is available via an analytic expression~\citep{diaconis1979conjugate}. An exponential family distribution $p(\vt{y})$ can be parameterized by a vector of expectation parameters $\vt{\mu} = \mean{\vt{T}(\vt{y})}{p(\vt{y})}$, where $\vt{T}(\vt{y})$ is the vector of sufficient statistics of the data. We can assign to these parameters a conjugate prior distribution $p(\vt{\mu})$, which in turn is parameterized by the prior expectation $\bar{\vt{\mu}}_0 = \mean{\vt{\mu}}{p(\vt{\mu})}$. Upon observing $N$ independently sampled datapoints, under some boundary assumptions, it can be shown that the posterior expectation parameters are a convex combination of the prior parameters $\bar{\vt{\mu}}_0$ and the maximum likelihood estimator~\citep{diaconis1979conjugate}: 
\begin{equation}
    \bar{\vt{\mu}} = \vt{\lambda} \odot \bar{\vt{\mu}}_0 + (1 - \vt{\lambda}) \odot \vt{\mu}_{\text{ML}}~,
\end{equation}
where $\vt{\lambda}$ is a vector of convex combination coefficients and $\odot$ denotes the element-wise product. A derivation of this result is given in Supplementary Material~A.

For example, in a Gaussian model with known precision $\tau$ and a Gaussian prior over the mean, the posterior mean is given by
\begin{equation}
    \bar{\mu} = \frac{\tau_0}{\tau_0 + N \tau} \bar{\mu}_0 + \frac{N \tau}{\tau_0 + N \tau} \left(\frac{1}{N} \sum_{n=1}^N y_n \right)~,
\end{equation}
where $\tau_0$ is the precision of the prior. This formula shows that the posterior parameters are a trade-off between the prior hyper-parameters and the values induced by the data.

\section{Convex-update variational families}

\makeatletter
\lst@CCPutMacro
    \lst@ProcessOther {"2D}{\lst@ttfamily{-{}}{-}}
    \@empty\z@\@empty
\makeatother
\begin{figure*}

\centering
  \begin{minipage}{.48\linewidth}
    \begin{smallpython}
    import tensorflow as tf
    import tensorflow_probability as tfp
    tfd = tfp.distributions
    num_timesteps = 20
    innovation_scale = 0.1
    
    @tfd.JointDistributionCoroutineAutoBatched
    def brownian_motion():
      xt = 0.
      for t in range(num_timesteps):
        xt = yield tfd.Normal(loc=xt,
                              scale=innovation_scale)
    \end{smallpython}
  \end{minipage}
  \begin{minipage}{.48\linewidth}
    \begin{smallpython}
    # Omitted for brevity: definitions of variational
    # parameters `lam` and `alpha` as `tf.Variable`s.
    @tfd.JointDistributionCoroutineAutoBatched
    def brownian_motion_surrogate_posterior():
      xt = 0.
      for t in range(num_timesteps):
        xt = yield tfd.Normal(
          loc=(lam[t]['loc'] * xt +
               (1 - lam[t]['loc']) * alpha[t]['loc']),
          scale=(lam[t]['scale'] * innovation_scale + 
                 (1 - lam[t]['scale']) *
                  alpha[t]['scale']))
    \end{smallpython}
  \end{minipage}
  \caption{TensorFlow Probability models for a discretized Brownian motion prior (left) and corresponding structured variational surrogate (right). In TFP's coroutine syntax, \protect\v{yield}ing a distribution samples a new random variable. The surrogate program is a written-out version of the process implicitly (and automatically) defined by the automatic method in Figure~\ref{fig:asvi_pseudocode}. Note that it has the same control-flow structure as the original model.}
\end{figure*}

Consider the following probabilistic model:
\begin{equation}
    p(x, y) = L(y\mid x) \rho(x\mid \vt{\theta})~,
\end{equation}
where $L\left(y\mid x\right)$ is a likelihood function and $\rho\left(x\mid\vt{\theta}\right)$ is a prior distribution parameterized by a vector of parameters $\vt{\theta}$. We do not assume the likelihood or the prior to be in the exponential family; simply that the parameters $\vt{\theta}$ are defined in a convex domain. We can construct a convex-update parameterized variational family by mimicking the form of the parameter update rule in conjugate models:
\begin{equation} \label{eqn:parameter update rule}
    q (x ; \vt{\lambda}, \vt{\alpha}) = \rho(x\mid \vt{\lambda} \odot \vt{\theta} + (1 - \vt{\lambda}) \odot \vt{\alpha})~,
\end{equation}
where $\vt{\lambda}$ is now a vector of learnable parameters with entries ranging from $0$ to $1$ and $\vt{\alpha}$ is a vector of learnable parameters with the same range of possible values as $\vt{\theta}$. In practice, it is convenient to express each $\lambda_i$ as the logistic sigmoid of a logit parameter, so that we can perform unconstrained optimization in the logit space.

In a model with a single latent variable, the convex-update parameterization is over-parameterized and equivalent to mean field. However, the power of this approach becomes evident in multivariate models constructed by chaining basic probability distributions together. Consider a probabilistic program specified in the form of Eq.~\eqref{eq: probabilistic program}. We can construct a structured variational family by applying the convex-update form to each latent conditional distribution in the model:
\begin{align} \label{eq:variational family}
    &q(\vt{x}; \vt{\Lambda}, \vt{A}) =  \prod_j \rho_j\!\left(x_j\mid \mathcal{U}_{\vt{\lambda}_j}^{\vt{\alpha}_j}\left[\vt{\theta}_j(\vt{\pi}_j) \right] \right)
\end{align}
with $\vt{\Lambda} = (\vt{\lambda}_1,\dots,\vt{\lambda}_J)$, $\vt{A} = (\vt{\alpha}_1, \dots, \vt{\alpha}_J)$ and the \emph{convex update operator}:
\begin{equation} \label{eq: convex update operator}
    \mathcal{U}_{\vt{\lambda}}^{\vt{\alpha}}\left[\vt{\theta} \right] = \vt{\lambda} \odot \vt{\theta} + (1 - \vt{\lambda}) \odot \vt{\alpha}~.
\end{equation}

The multivariate structured distributions induced by this family have several appealing theoretical properties that justify their usage in structured inference problems:
\begin{enumerate}
    \item The family always contains the original probabilistic program (i.e. the prior distribution). This is trivial to see as we can obtain the prior by setting $\vt{\lambda} = \vt{1}$.  On the other hand, setting $\vt{\lambda} = \vt{0}$ results in the standard mean-field approximation. Note that none of the commonly used automatic structured variational approaches share this basic property. 
    
    \item The family includes both the exact filtering and smoothing posterior of univariate linear Gaussian time series models of the form
    \begin{equation}
        x_{t} \sim \mathcal{N}\! \left( a x_{t-1}, \sigma^2 \right), ~~~ y_t \sim \mathcal{N}\! \left( x_{t}, \xi^2 \right)~.
    \end{equation}
    In this case, the filtering conditional posterior is given by the Kalman filter update: 
    \begin{align} \label{eq: kalman update}
        \bar{\mu}_{t+1} &= (1 - K_t) {\mu}_{t+1}(x_t) + K_t y_t \\ \nonumber
        &= (1 - K_t) a x_{t} + K_t y_t~,
     \end{align}
     where $0 \leq K_t \leq 1$ is the Kalman gain, corresponding to $\lambda_t$ in ASVI. The smoothing update has a similar form where the data term is augmented with an estimate integrating all future observations. 
     
     \item The convex-update family has a very parsimonious parameterization compared to other structured families. The number of parameters is $2 P$, where $P$ is the total number of parameters of the conditional distributions. In contrast, the common multivariate normal approach scales as $P^2$. However, this parsimonious parameterization implies that the convex-update family cannot capture dependencies that are not already present in the prior of the probabilistic program. Specifically, the convex-update family cannot model correlations originating from colliding arrows in the directed acyclic graph (`explaining away' dependencies).
\end{enumerate}

\section{Discrete latent variables and stochastic control flow}
While this paper is focused on differentiable probabilistic programming, the pseudo-conjugate family can be applied equally well to models with discrete latent variables and combinatorial link functions. Consider a conditional Bernoulli variable:
\begin{equation}
    p(b \mid \vt{\pi}) = \rho(\vt{\pi})^b (1 - \rho(\vt{\pi}))^{1 - b}~,
\end{equation}
where the array $\vt{\pi}$ collects the values of its parents. Since the parameter $\rho$ is defined in a convex domain, we can construct a convex-update variational posterior:
\begin{equation}
    p(b \mid \vt{\pi}) =  \mathcal{U}_{\vt{\lambda}}^{\vt{\alpha}}\left[\vt{\rho(\vt{\pi})} \right]^b (1 - \mathcal{U}_{\vt{\lambda}}^{\vt{\alpha}}\left[\vt{\rho(\vt{\pi})} \right]))^{1 - b}~.
\end{equation}
It is easy to see that the same procedure can be applied to binomial, categorical, multinomial and Poisson variables among others. Using discrete variables, we can also construct convex-update families with stochastic control flows. For example, a binary latent $b$ can determine whether the stochastic computation proceeds on one branch of the computation tree rather than another. This has the nice feature of having an approximate posterior program with the same stochastic flow structure of the prior program where the probability of each gate is biased by the observed data.

\section{Automatic structured variational inference}

\begin{figure}[t]
\begin{smallpython}
def build_asvi_surrogate_posterior(prior):
  ConvexUpdateParams = namedtuple(['lam', 'alpha'])
  q_vars = {}
  @tfd.JointDistributionCoroutineAutoBatched
  def asvi_surrogate_posterior():
    # Step the model to yield the first RV.
    prior_gen = prior._coroutine_fn()
    rv = next(prior_gen)
    while True:  # Run model to termination.
      # If this is a new RV, initialize variables.
      if rv.name not in q_vars:
        q_vars[rv.name] = {
          k: ConvexUpdateParams(
            lam=Variable(random_init_like(v),
                         xform=Sigmoid()),
            alpha=Variable(random_init_like(v),
                           xform=dist.bijectors[k]()))
          for k, v in rv.parameters.items()}
      # Apply convex update for each parameter.
      q_params = {}
      for k, v in rv.parameters.items(): 
        lam = q_vars[rv.name][k].lam
        alpha = q_vars[rv.name][k].alpha
        q_params[k] = lam * v + (1. - lam) * alpha
      # Step to the next RV.
      q_sample = yield type(rv)(**q_params)
      rv = prior_gen.send(q_sample)
  return asvi_surrogate_posterior
\end{smallpython}
\caption{Python pseudocode for general-purpose automated construction of a structured variational family. A full implementation is provided in TensorFlow Probability (\href{https://www.tensorflow.org/probability/api_docs/python/tfp/experimental/vi/build_asvi_surrogate_posterior}{\protect\v{tfp.experimental.build_asvi_surrogate_posterior}}).}
\label{fig:asvi_pseudocode}
\end{figure}

Construction of the variational family (\ref{eq:variational family}) can be straightforwardly automated by probabilistic programming systems. A variational program is defined by running the input program under a nonstandard interpretation: at each random variable site 
\[
x^\text{prior}_j \sim p_j\left(x_j \mid \vt{\theta}_j\left((\vt{\pi}^\text{prior}_j\right)\right)\,,
\] 
the variational program instead samples from the convex-update distribution (Eq.~\eqref{eqn:parameter update rule}), 
\[
x^\text{posterior}_j \sim q\left(x_j \mid \vt{\theta}_j\left(\vt{\pi}^\text{posterior}_j\right); \vt{\lambda}_j, \vt{\alpha}_j\right)\,,
\] where $\vt{\lambda}_j$ and $\vt{\alpha}_j$ are trainable variables. Because each variable definition is rewritten `locally', the variational program has the same control-flow structure, dependence graph, and time complexity as the original model. Local transformations of this kind may be implemented by effect handlers~\citep{plotkin2009handlers}, a mechanism supported in multiple recent probabilistic programming frameworks~\citep{moore2018effect,bingham2019pyro,phan2019composable}. 

We provide an open-source implementation of ASVI for joint distribution models in TensorFlow Probability~\citep[TFP; ][]{dillon2017tensorflow}, which uses an effect-handling-like mechanism (Python coroutines) to transform an input joint distribution into a structured variational surrogate. Figure~\ref{fig:asvi_pseudocode} shows a simplified version.

Our implementation operates in the default parameterization $\vt{\theta}$ of each TFP distribution class, which is typically similar (if not exactly equal) to the expectation parameterization; e.g., normal distributions expose a location-scale parameterization. Constrained variational parameters $(\vt{\lambda}_j, \vt{\alpha}_j)$ are defined as differentiable transformations of unconstrained variables, where each distribution is annotated with appropriate constraining transformations for its parameters. For example, the softplus transformation $f(x) = \log(1 + \exp(x))$ enforces that scale parameters have positive values. For variables such as log-normal distributions whose definition involves a bijective transformation $\vt{y} = f_\psi(\vt{x})$ of a base variable $\vt{x} \sim p_\phi(\vt{x})$, the convex-update family has an analogous form and defines a convex update on the union of the parameters $\vt{\theta} = (\vt{\phi}, \vt{\psi})$.

Given a structured variational program, the ELBO (\ref{eqn:elbo}) may be estimated by Monte Carlo sampling and optimized with stochastic gradient estimates, obtained via the reparameterization trick for continuous-valued variables and score-function estimator for discrete variables~\citep{ranganath2014black,kucukelbir2017automatic}.

\section{Related work}

Most structured VI approaches require model specific derivations and variational bounds. However, several forms of model-agnostic structured variational distributions have been introduced. The most commonly used fully automatic approach is probably automatic differentiation variational inference (ADVI) ~\citep{kucukelbir2017automatic}. The ADVI variational family is constructed by mapping the values of all latent variables to an unbounded coordinate space based on the support of each distribution. The variational distribution in this new space is then parameterized as either a spherical (mean field) or a fully multivariate normal distribution. While this approach is broadly applicable, it exploits very little information from the original probabilistic model and has scalability problems due to the cubic complexity of Bayesian inference when the multivariate distribution is used. Hierarchical VI accounts for dependencies between latent variables by coupling the parameters of their factorized distributions through a joint \emph{variational prior}~\citep{ranganath2016hierarchical}. While this method is very generic, it requires user input in order to define the variational prior and it uses a modified variational lower bound. Copula VI models the dependencies between latent variables using a vine copula function~\citep{tran2015copula}. In the context of probabilistic programming, copula VI shares some of the same limitations of hierarchical VI: it requires the appropriate specification of bivariate copulas and it needs a specialized inference technique. The approach that is closest to our current work is structured stochastic VI~\citep{hoffman2013stochastic}. Similar to our approach, its variational posteriors have the same conditional independence structure as the input probabilistic program. However, this method is limited to conditionally conjugate models with exponential family distributions. Furthermore, the resulting ELBO is intractable and needs to be estimated using specialized techniques. Similarly, conjugate-Computation Variational Inference uses conjugate Bayesian updates in non-conjugate models with locally conjugate components~\citep{khan2017conjugate}. This differs from our approach as we use the convex updates for all the variables in the model. 

Normalizing flows are general deep learning methods designed to approximate arbitrary probability densities through a series of learnable invertible mappings with tractable Jacobians~\citep{rezende2015variational,kingma2016improved,dinh2016density,kingma2018glow,papamakarios2017masked, kobyzev2020normalizing}. Flows have been used in stochastic VI as very expressive and highly parameterized variational families~\citep{rezende2015variational, kingma2016improved}. Most normalizing flow architectures used in VI do not incorporate the conditional independence structure of the prior probabilistic program. Structured conditional continuous normalizing flows are a new class of normalizing flows that have the conditional independence structure of the true posterior~\citep{weilbach2020structured}. These architectures are based on faithful inversion~\citep{webb2018faithful} and implement the dependency structure using sparse matrices. Conversely to ASVI, this approach only exploits the graphical structure of the probabilistic program and therefore ignores the specific form of the conditional distributions and link functions.

Structured VI is most commonly applied in time series models such as hidden Markov models and autoregressive models. In these cases the posterior distributions inherit strong statistical dependencies from the sequential nature of the prior. Structured VI for time series usually use structured variational families that capture the temporal dependencies, while being fully-factorized in the non-temporal variables~\citep{eddy1996hidden, foti2014stochastic, johnson2014stochastic, karl2016deep, fortunato2017bayesian}. This differs from our convex-update families, where both temporal and non-temporal dependencies are preserved. 

\section{Applications}
We first evaluate ASVI on a set of standardized Bayesian inference tasks to compare ASVI to other automated inference methods. We then apply ASVI to a both a deep Bayesian smoothing and a deep generative modeling task, to demonstrate that ASVI scales to large datasets while producing high quality posteriors. The code needed to run ASVI and baselines on the standardized tasks can be found in \href{https://github.com/google-research/google-research/tree/master/automatic_structured_vi}{this repository}.

\subsection{Inference Gym tasks}

\begin{table*}[p]
\footnotesize
\centering
\caption{Average final negative ELBO values and their standard errors, computed over $15$ simulations. Each row corresponds to an Inference Gym task and each column is a posterior baseline. Boldface indicates the best performance.}
\begin{tabular}{lcccccc}
\centering
\small
    {} & {ASVI} & {Mean field} & {Small IAF} & {Large IAF} & {MVN} & {AR(1)} \\ \midrule
    {BR}      & $\boldsymbol{-5.20 \pm 0.11}$      & $0.99 \pm 0.49$    & $-3.47 \pm 0.24$      & $-4.9 \pm 0.12$       & $-3.7 \pm 0.2$          & $-4.01 \pm 0.31$ \\
    {BRG}     & $-0.35 \pm 0.10$     & $-5.16 \pm 0.31$    & $-1.31 \pm 0.23$     & $\boldsymbol{0.14 \pm 0.09}$      & $-2.27 \pm 0.20$       & $-2.08 \pm 0.31$ \\
    {LZ}      & $\boldsymbol{34.29 \pm 0.66}$    & $1225.66 \pm 4.30$ & $1236.49 \pm 12.47$ & $1242.11 \pm 10.39$ & $2153.08 \pm 179.24$  & $1271.36 \pm 16.53$\\
    {LZG}     & $\boldsymbol{34.68 \pm 0.15}$    & $119.67 \pm 0.17$  &  $112.64 \pm 0.58$  & $94.8 \pm 0.14$     & $98.43 \pm 0.09$      & $54.53 \pm 2.1$\\
    {ES}      & $36.50 \pm  0.04$   & $36.94 \pm  0.04$  & $36.28 \pm 0.04$    & $\boldsymbol{36.26 \pm  0.02}$   & $36.58 \pm  0.02$   & N/A\\
    {R}       & $1079.8 \pm 0.22$  & $1082.77 \pm 0.59$ & $1078.85 \pm 0.21$  & $\boldsymbol{1078.52 \pm 0.19}$  & $1090.96 \pm 0.48$  & N/A \\
\end{tabular}
\label{tab:ig-elbos}
\end{table*}

\begin{table*}[p]
\footnotesize
\centering
\caption{Mean and standard error of the absolute error in posterior mean (M) and standard deviation (SD) parameters relative to MCMC ground truth, normalized by true SD, computed over $15$ simulations. Boldface indicates the best performance.}
\begin{tabular}{llcccccc}
\centering
\small
{} & {} & {ASVI} & {Mean field} & {Small IAF} & {Large IAF} & {MVN} & {AR(1)} \\ \midrule
{BR}      & M & $0.16 \pm 0.03$ & $0.14 \pm 0.02$ & $0.1 \pm 0.02$ & $0.12 \pm 0.03$ & $\boldsymbol{0.09 \pm 0.01}$ & $0.24 \pm 0.05$ \\
& SD & 
$\boldsymbol{0.06 \pm 0.02}$ &  $0.37 \pm 0.01$ & $0.14 \pm 0.03$ &  $0.07 \pm 0.02$ & $0.07 \pm 0.01$ &  $0.07 \pm 0.01$  \\ \midrule
{BRG}     & M & $\boldsymbol{0.69 \pm 0.07}$ &
     $0.81 \pm 0.03$ &
     $0.77 \pm 0.03$ &
     $0.71 \pm 0.05$ &
     $0.74 \pm 0.03$ &
     $0.80 \pm 0.07$ \\
& SD &
     $\boldsymbol{0.22 \pm 0.02}$ &
     $0.31 \pm 0.01$ &
     $0.27 \pm 0.02$ &
     $0.23 \pm 0.02$ &
     $0.23 \pm 0.02$ &
     $0.24 \pm 0.02$ \\ \midrule
{LZ}      & M & $\boldsymbol{0.36 \pm 0.25}$ &
     $35.83 \pm 0.00$ &
     $35.94 \pm 0.11$ &
     $36.62 \pm 0.07$ &
     $39.66 \pm 0.00$ &
     $36.52 \pm 0.18$ \\
& SD &
     $\boldsymbol{0.47 \pm 0.03}$ &
     $0.94 \pm 0.00$ &
     $0.92 \pm 0.01$ &
     $0.81 \pm 0.01$ &
     $0.67 \pm 0.01$ &
     $0.83 \pm 0.01$ \\ \midrule
{LZG}     & M &
     $\boldsymbol{0.15 \pm 0.03}$ &
     $21.46 \pm 0.04$ &
     $23.35 \pm 0.16$ &
     $23.96 \pm 0.06$ &
     $23.86 \pm 0.08$ &
     $1.56 \pm 0.24$ \\
& SD &
     $\boldsymbol{0.39 \pm 0.02}$ &
     $1.51 \pm 0.04$ &
     $1.51 \pm 0.20$ &
     $0.80 \pm 0.14$ &
     $2.59 \pm 0.06$ &
     $2.59 \pm 0.08$ \\ \midrule
{ES}      & M & $0.16 \pm 0.03$ &
     $0.16 \pm 0.02$ &
     $0.16 \pm 0.04$ &
     $\boldsymbol{0.13 \pm 0.04}$ &
     $\boldsymbol{0.13 \pm 0.04}$ &
     N/A  \\
& SD &
     $0.07 \pm 0.02$ &
     $\boldsymbol{0.05 \pm 0.01}$ &
     $0.07 \pm 0.02$ &
     $0.08 \pm 0.03$ &
     $0.06 \pm 0.01$ &
     N/A  \\ \midrule
{R} & M & $\boldsymbol{0.12 \pm 0.02}$ &
     $0.24 \pm 0.02$ &
     $0.17 \pm 0.02$ &
     $0.15 \pm 0.02$ &
     $0.42 \pm 0.01$ &
     N/A \\
& SD &
     $0.15 \pm 0.03$ &
     $0.12 \pm 0.01$ &
     $\boldsymbol{0.09 \pm 0.01}$ &
     $0.10 \pm 0.02$ &
     $0.50 \pm 0.02$ &
     N/A \\ \midrule
\end{tabular}
\label{tab:ig-mean-std}
\end{table*}

Inference Gym~\citep[IG; ][]{inferencegym2020} is a Python library for evaluating probabilistic inference algorithms. It defines a set of standardized inference tasks and provides implementations as TFP joint distribution models~\citep{piponi2020joint} and as Stan programs~\citep{carpenter2017stan}, which are used to compute `ground truth' posterior statistics via MCMC.

For each task we evaluate ASVI against a suite of baseline variational posteriors, including mean field (MF), inverse autoregressive flows~\citep{kingma2016improved, papamakarios2017masked} with eight hidden units (Small IAF) and 512 hidden units (Large IAF), and a multivariate normal posterior (MVN). Where possible, we also compare to an AR(1) posterior. For each method the ELBO is optimized to convergence (up to 100000 steps, although significantly fewer were required in most cases) using the Adam optimizer, with learning rate set by a hyperparameter sweep for each task/method pair---this tuning was particularly important for the IAF models. Further details are provided in the supplement.


{\bf Time series.} We consider two discretized stochastic differential equation (SDE) models, defined by conditional densities of the following form: 
\begin{align}
    x_{t+1} &\sim \N\left(x_t + f(x_{t}, t) \dt, g(x_t, t)^2 \dt \right) \\
    y_t &\sim \N\left(x_t, \sigma^2_\text{obs}\right)
\end{align}
where $x$ represents the latent process with drift function $f$ and volatility function $g$, and observations $y$ are taken with noise standard deviation $\sigma_\text{obs}$. The first model describes Brownian (BR) motion without drift, and the second model is a stochastic Lorenz dynamical system (LZ). Full model specifications are provided in the supplementary material. We also include variants of both models that include global variables (BRG and LZG) where the  innovation and observation noise scale parameters are unknown. We simulate each model forward 30 steps and assume that noisy observations are taken of the first ten and last ten steps, and the middle ten are unobserved.




{\bf Hierarchical regression models.} We also evaluate ASVI on two standard Bayesian hierarchical models: Eight Schools~\citep[ES; ][]{gelman2013}, which models the effect of coaching programs on standardized test scores, and Radon~\citep[R; ][]{gelman2007}, a hierarchical linear regression model that predicts Radon measurements taken in houses. Full descriptions are provided in the supplementary material.

\begin{figure}[tb]
    \centering
    \includegraphics[width=0.45\textwidth]{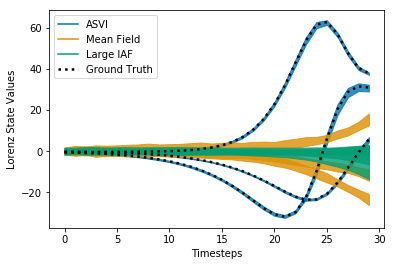}
    \caption{Qualitative results of the Lorenz system with globals (LZG) experiment, showing each baseline's estimate of the three latent dimensions in the Lorenz model $\pm$ two standard deviations. Only ASVI is able to successfully infer the values of each latent series.}
    \label{fig:lorenzbridgeresults}
\end{figure}

{\bf Results.} \autoref{tab:ig-elbos} shows that ASVI and the large IAF produce the best ELBOs; ASVI is competitive with the large IAF on all tasks despite having far fewer parameters. Similarly, in \autoref{tab:ig-mean-std} we see that ASVI's estimates of posterior means and standard deviations are consistently among the best relative to MCMC ground truth. Mean field fails to converge to the true posterior estimate in all four models, indicating that inclusion of prior structure in ASVI is needed to capture the dependencies in the true posterior. Qualitatively, we observe in \autoref{fig:lorenzbridgeresults} that only the ASVI posterior captures the ground truth posterior in the Lorenz task, indicating that ASVI is able to capitalize on strong prior structure when it exists. Figure 1 in the supplementary material shows loss trajectories. For all but the Radon model, ASVI tends to converge in fewer iterations than the large IAF, the other highest-performing surrogate posterior.

\begin{figure}
    \centering
    \includegraphics[width=0.5\textwidth]{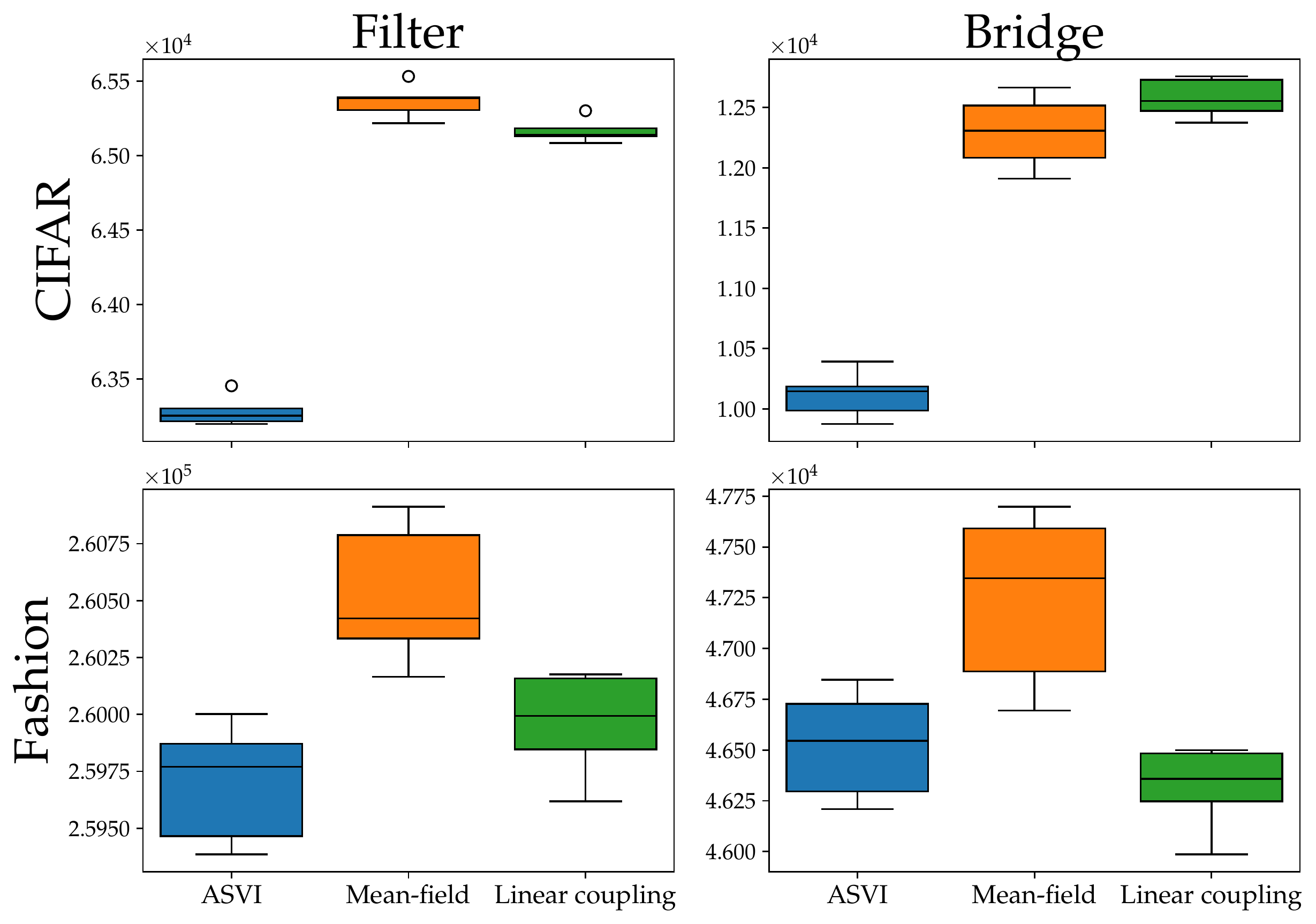}
    \caption{Quantitative results of the deep Bayesian smoothing experiment in \autoref{sec:deep-bayesian-smoothing}. Shown is the negative ELBO of ASVI, mean-field, and linear coupling at the final training iteration.}
    \label{fig:deepbayesianlosses}
    \vspace{-1em}
\end{figure}

\begin{figure*}
    \centering
    \includegraphics[width=\textwidth]{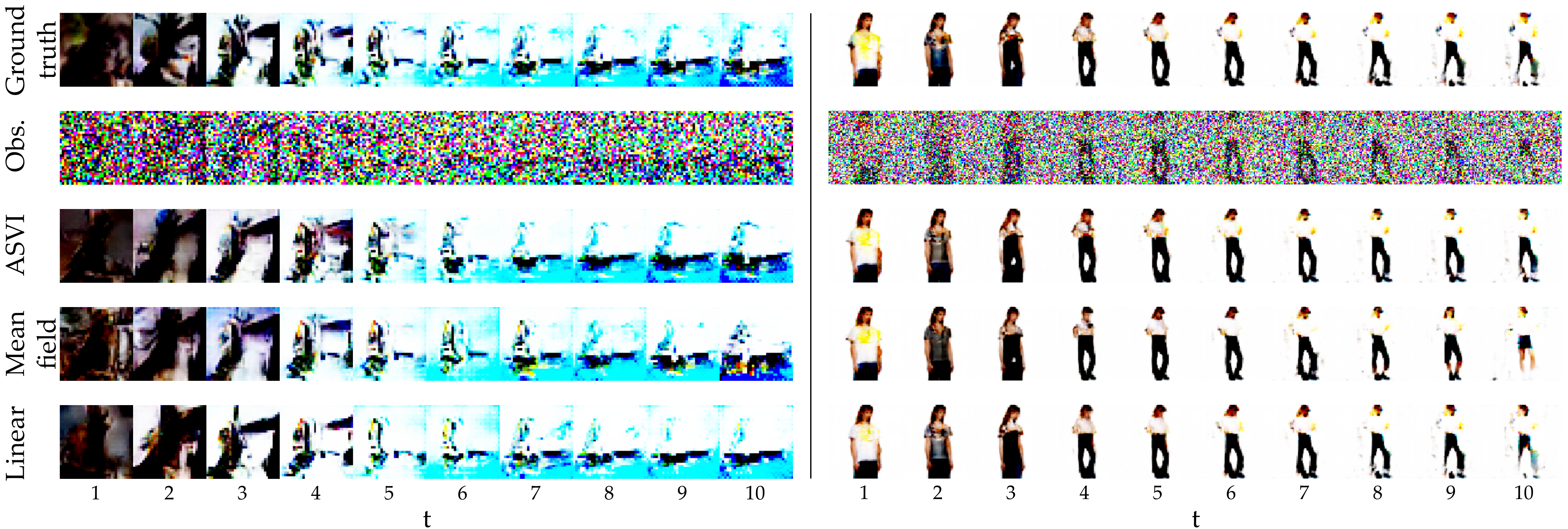}
    \caption{Qualitative results of the deep Bayesian filters using ASVI, mean-field, and linear coupling for CIFAR10 (left) and FashionGEN (right). The top row shows the ground truth images, the second row the noisy-corrupted observations, and the remaining rows show the different reconstructions.}
    \label{fig:deepbayesianfilter}
\end{figure*}

\subsection{Deep Bayesian smoothing with neural SDEs}
\label{sec:deep-bayesian-smoothing}
So far, we performed inference in simple time series and hierarchical models with relatively low-dimensional state spaces. We will now test the performance of ASVI on a high-dimensional problem with complex nonlinearities parameterized by deep networks. As a latent model, we use neural (discretized) stochastic differential equations (SDE)~\citep{chen2018neural, li2020scalable}:
\begin{equation}
    \text{d} \vt{x} = \vt{F}(\vt{x}) \text{d}t + \text{d} \vt{B}(t)
\end{equation}
where $\vt{F}(\vt{x})$ is a nonlinear function parameterized by a neural network and $\vt{B}(t)$ is standard multivariate Brownian motion (see Supplementary Material~C for the details of the architecture). This latent process generates noise-corrupted observations through a deep network $\vt{G}$:
\begin{equation}
    \vt{y}_t \sim \mathcal{N}\!\left(\vt{G}(\vt{x}(t)), \sigma^2 \right) \,.
\end{equation}
In our examples, $\vt{G}$ is a generator which converts latent vectors into RGB images. We tested two kinds of pre-trained generators: A DCGAN trained on CIFAR10 and a DCGAN trained on FashionGEN~\citep{radford2015unsupervised} (see Supplementary Material~C). In the former case, the latent space is $100$-dimensional, while in the latter it is $120$-dimensional. We considered two inference challenges: filtering, where we aim to remove the noise from a series of images generated by a trajectory in the latent process, and bridging, where we reconstruct a series of intermediate observations given the beginning (first three time points) and the end (last two time points) of a trajectory. In both cases, we assume knowledge of the dynamical and generative models, discretize the neural SDE using an Euler–Maruyama scheme and backpropagate through the integrator~\citep{chen2018neural}.

Figure~\ref{fig:deepbayesianfilter} shows the filtering performance of ASVI and two baselines (mean field, and linear Gaussian model; see Supplementary Material~C for the details) in a bridging problem. The quantitative results (negative ELBOs) are shown in Figure~\ref{fig:deepbayesianlosses}. ASVI always reaches tighter lower bounds except in the Fashion bridge experiment where it has slightly lower performance than the linear coupling baseline. This tighter variational bound results in discernibly higher quality filtered images in both CIFAR10 and Fashion, as shown in Figure~\ref{fig:deepbayesianfilter}. The figure also shows several samples from the ASVI bridge posterior. As expected, the generated images diverge in the unobserved period and reconverge at the end.

\subsection{Deep amortized generative modeling}



\begin{figure*}[htb]
    \centering
    \includegraphics[width=\textwidth]{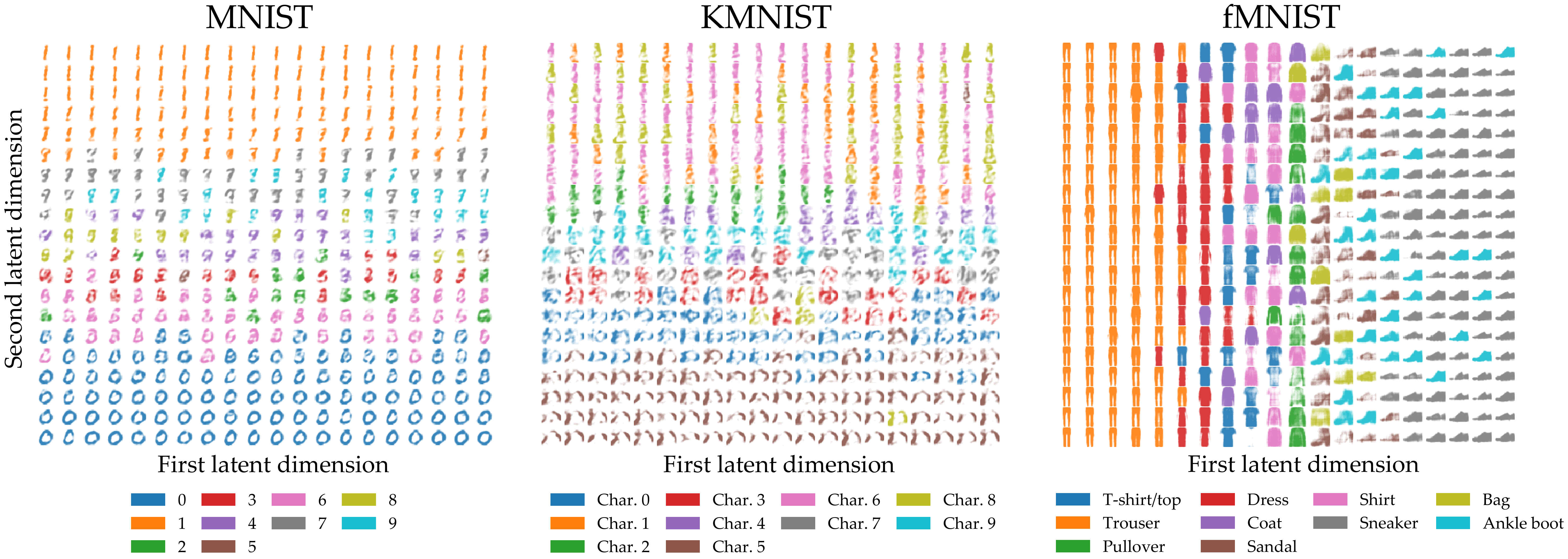}
    \caption{The latent embeddings for the different MNIST variants. Each element is a predicted sample together with its corresponding label.}
    \label{fig:fMNIST_embedding}
\end{figure*}
\begin{table*}[ht] 
\centering
\small
\caption{Average and standard error of the RMSE between the posterior means and the ground truth functions, computed over $15$ simulations. Boldface indicates the best performance.}
\begin{tabular}{lrrrrrr}
\toprule
& \multicolumn{2}{c}{MNIST} & \multicolumn{2}{c}{kMNIST} & \multicolumn{2}{c}{fMNIST}   \\
\cmidrule(lr){2-3}\cmidrule(lr){4-5}\cmidrule(lr){2-3}\cmidrule(lr){6-7}
         & \multicolumn{1}{c}{no labels}      & \multicolumn{1}{c}{labels}     & \multicolumn{1}{c}{no labels} & \multicolumn{1}{c}{labels} & \multicolumn{1}{c}{no labels} & \multicolumn{1}{c}{labels} \\ \midrule
ASVI      & $\boldsymbol{-134.6} \pm 0.1$ & $\boldsymbol{-99.7} \pm 0.5$ & $\boldsymbol{-252.4} \pm 0.3$ & $\boldsymbol{-200.3} \pm 0.7$   & $\boldsymbol{-151.1} \pm 0.4$ & $\boldsymbol{-116.4} \pm 1.2$   \\ 
MF & $-153.8 \pm 0.2$ & $-146.3 \pm 0.5$ & $-267.0 \pm 0.3$ & $-258.6 \pm 0.8$   & $-166.7 \pm 0.2$ & $-158.2 \pm 1.3$   \\ 
\bottomrule
\end{tabular} 
\label{tab: VAE}
\end{table*}

Finally, we apply an amortized form of ASVI to a deep generative modeling problem. The goal is to model the joint distribution of a set of binary images $\vt{y}$ paired with class labels $l$. We use a deep variational autoencoder with three layers of latent variables $\vt{z}_1$, $\vt{z}_2$ and $\vt{z}_3$ coupled in a feed-forward fashion through ReLU fully-connected networks: 
\begin{equation}
    \begin{split}
        \vt{z}_1 &\sim \mathcal{N}\!\left( 0, 1.5 \right) \\
        \vt{z}_2 &\sim \mathcal{N}\!\left( \text{ReLU}\left(\vt{f}_1(\vt{z}_1)\right), 0.25 \right) \\ 
        \vt{z}_3 &\sim \mathcal{N}\!\left( \text{ReLU}\left(\vt{f}_2(\vt{z}_2)\right), 0.1 \right) \\ 
        \vt{y} &\sim \mathcal{N}\!\left( \vt{g}_1(\vt{z}_3), 0.1 \right) \\
        l &\sim \text{Categorical}\!\left( \text{Softmax}\!\left(\vt{g}_2(\vt{z}_2)\right)\right) 
    \end{split}
\end{equation}
where $\vt{f}_1$ and $\vt{f}_2$ are fully-connected two-layer networks with ReLU activations, linear output units, and $25$ and $75$ hidden units respectively,  while $\vt{g}_1$ and $\vt{g}_2$ are linear layers. The amortized convex-update distribution has the following form:
\begin{equation} \label{eq: amortized distribution}
    \begin{split}
        \vt{z}_1 &\sim \mathcal{N}\!\left(\vt{\alpha}_1(\vt{y}), \vt{\xi}_1(\vt{y}) \right) \\ 
        \vt{z}_2 &\sim \mathcal{N}\!\left(\vt{\lambda}_2 \vt{f}_1(\vt{z}_1) + (1 - \vt{\lambda}_2) \vt{\alpha}_2(\vt{y}), \vt{\xi}_2(\vt{y}) \right) \\
        \vt{z}_3 &\sim \mathcal{N}\!\left(\vt{\lambda}_3 \vt{f}_2(\vt{z}_1) + (1 - \vt{\lambda}_3) \vt{\alpha}_3(\vt{y}), \vt{\xi}_3(\vt{y}) \right)
    \end{split}
\end{equation}
where the mean vectors $\vt{\alpha}_k(\vt{y})$ are the activation of the $(8 - 2k)$-th layer (post ReLU) of a fully-connected 6-layer ReLU inference network (with layer sizes $(120, 100, 70, 50, 70, 2)$) that takes the image $\vt{y}$ as input. The scale parameter vectors $\vt{\xi}_k(\vt{y})$ were obtained by connecting a linear layer to the $(8 - 2k)$-th layer, followed by a softplus transformation. The details of all architectures are given in Supplementary Material~D. The amortized family was parameterized by $\vt{\lambda}$ and the weights and biases of the inference network. Note that this amortization scheme is not fully automatic as it require the choice of the inference networks. The mean-field baseline had the same form given in Eq.~\eqref{eq: amortized distribution}, but with the expectation of the distribution fully determined by $\vt{\alpha}_k(\vt{y})$. We did not include a comparison with the other baselines as it is computationally unfeasible in this larger scale experiment.

We tested the performance of these deep variational generative models on three computer vision datasets: MNIST, FashionMNIST and KMNIST~\citep{lecun1998gradient, xiao2017fashion, clanuwat2018deep}. Furthermore, we performed two types of experiment. In the first, images and labels were generated jointly. In the second, only images were generated. Table~\ref{tab: VAE} reports the performance of ASVI and mean-field baseline quantified as the test set ELBO. The ELBO was estimated by sampling the latent variables $20$ times per test image from the variational posterior. Mean ELBO and SEM were obtained by repeating the estimation $20$ times. ASVI achieves higher performance in all experiments for all datasets. Figure~\ref{fig:fMNIST_embedding} shows a randomized selection of images generated by the ASVI model together with the corresponding label.

\section{Discussion}
We introduced an automatic algorithm for constructing an appropriately structured variational family, given an input probabilistic program. The resulting method can be used on any probabilistic program specified by a directed Bayesian network, and always preserves the forward-pass structure of the input program. The main limitation of the convex-update family is that it cannot capture dependencies induced by colliding arrows in the input graphical model. Consequently, in a model such as a standard Bayesian neural network, where the prior over the weights is decoupled, the convex-update family is a mean-field family.
Despite this limitation, our results demonstrate good performance (competitive with a high-capacity IAF) on small Bayesian hierarchical models with colliding arrows in the prior graphical model, through capturing the forward structure alone. Designing structured surrogates that efficiently capture the full posterior dependence structure is an interesting direction for ongoing work.  

\clearpage
\bibliographystyle{unsrtnat}
\bibliography{ref.bib}

\appendix

\onecolumn

\section{Convex updates in conjugate models}
Consider an exponential family likelihood and its expectation parameter:
\begin{equation}
\mu(\eta) = \int e^{\eta \cdot T(x) - A(\eta) } T(x) \text{d} x = \nabla_\eta A(\eta)~,
\end{equation}
where $\eta$ is the natural parameter. The conjugate prior is
\begin{equation}
p(\eta\mid \tau_0,n_0) = h(\tau, n_0) e^{\tau_0 \cdot \eta - n_0 A(\eta)}~.
\end{equation}
where $h(\cdot)$ is a base density, $\tau_0$ is the relevant prior natural parameter and $n_0$ is the natural "prior count" parameter. This also induces a prior $p(\mu\mid \tau_0, n_0)$ over the expectation parameters $\mu$. In Section 2.1, we parameterize this prior $p(\mu\mid \tau_0, n_0)$ using the prior expectations of the expectation parameters
\begin{equation}
\bar{\mu}_0 = \int \mu p(\mu\mid \tau_0, n_0) \text{d} \mu = \int \nabla_\eta A(\eta) p(\eta\mid \tau_0, n_0) \text{d} \eta
\end{equation}
The result in Eq.~3 follows from the fact that $\bar{\mu}_0 = \tau_0/n_0$. To see this, we can integrate both sides of the trivial identity:
\begin{equation}
\nabla_\eta p(\eta \mid  \tau_0, n_0) = p(\eta \mid  \tau_0, n_0)(\tau_0 - n_0 \nabla_\eta A(\eta))
\end{equation}
and notice that $\int \nabla_\eta p(\eta \mid  \tau_0, n_0) \text{d} \eta = 0$. This result can be proven when $p(\eta \mid  \tau_0, n_0)$ vanishes at the boundary values of the integral using the generalized divergence theorem (Diaconis, 1979). Using this result, we can use the conjugate update rules $\tau = \tau_0 + \sum_n T(x_n)$ and $n = n_0 + N$ to obtain
\begin{align}
\bar{\mu} &= \frac{\tau}{n} = \frac{\tau_0 + \sum_n^N T(x_n)}{n_0 + N} \\
&= \left(\frac{n_0}{n_0 + N}\right) \frac{\tau_0}{n_0} + \left(1 - \frac{n_0}{n_0 + N}\right) \frac{1}{N} \sum_n^N T(x_n)\\
&= \lambda \bar{\mu}_0 + (1 - \lambda) \mu_{ML}
\end{align}
where $N$ is the number of datapoints and $\mu_{ML} = \frac{1}{N} \sum_n^N T(x_n)$ can be shown to be the maximum likelihood estimator of the expectation parameter.

\section{Details of the Inference Gym experiments}

\subsection{Inference Gym tasks}

\subsubsection*{Time series models}
As a first application, we focus on timeseries models and SDEs. We used two models, both from the Inference Gym~\citep{inferencegym2020}. 

The first model (BR) is a Brownian motion without drift, governed by $x'(t) = w_x(t)$, where $w_x(t)$ is a Gaussian white noise process with scale $\sigma_x$. The value of $x_t$ is observed with noise standard deviation equal to $\sigma_\text{obs}$. The data are generated with $\sigma_x=0.1$ and $\sigma_\text{obs}=0.15$. In the BRG model with global variables, both $\sigma_x$ and $\sigma_\text{obs}$ are treated as random variables with \v{LogNormal(loc=0, scale=2)} priors.

The second model (LZ) is a stochastic Lorenz system (nonlinear SDE):
\begin{subequations}
\begin{align}
    x'(t) &= 10 (y(t) - x(t)) + w_x(t)\\ 
    y'(t) &= x(t) (28 - z(t)) - y(t) + w_y(t) \\ 
    z'(t) &= x(t) y(t) - (8/3) z(t) + w_z(t) 
\end{align}
\end{subequations}
where $w_x(t)$, $w_y(t)$ and $w_z(t)$ are Gaussian white noise processes with standard deviation $\sigma = 0.1$. The value of $x(t)$ is observed with Gaussian noise with standard deviation $\sigma_\text{obs} = 1.$; $y(t)$ and $z'(t$ are left unobserved. When global variables are allowed (the LZG model), $\sigma$ and $\sigma_\text{obs}$ are treated as unknown random variables with \v{LogNormal(loc=-1., scale=1.)} priors.



All processes were discretized with the Euler–Maruyama method ($\dt = 0.01$ for BR and $dt = 0.02$ for LZ) and the transition probability was approximated as Gaussian (this approximation is exact for $dt$ tending to $0$). Each model was integrated for $30$ steps. 

\subsubsection*{Hierarchical models}

Eight Schools~\citep{gelman2013} models the effect of coaching programs on standardized test scores, and is specified as follows:
\begin{align}
    \mu &\sim \N( 0, 100 ) \\
    \log \tau &\sim \log \N( 5, 1) \\
    \theta_i &\sim \N(\mu, \tau^2) \\
    y_i &\sim \N (\theta_i, \sigma_i^2)
\end{align}
where $i = 1,\ldots,8$ indexes the schools, $\mu$ represents the prior average treatment effect and $\tau$ controls the variance between schools. The $y_i$ and $\sigma_i$ are observed.

The Radon model~\citep{gelman2007} is a Bayesian hierarchical linear regression model that predicts measurements of Radon, a carcinogenic gas, taken in houses in the United States. The hierarchical structure is reflected in the grouping of houses by county, and the model is specified as follows:
\begin{align}
    \mu &\sim \N(0, 1) \\
    \tau &\sim \N^+ (0, 1) \\
    \theta_i &\sim \N(\mu, \tau^2) \\
    \beta_1, \beta_2, \beta_3 &\sim \N(0, 1) \\
    \sigma &\sim \N^+ (0, 1) \\
    y_j &\sim \N(\beta_1z_{c_j} + \beta_2x_j + \beta_3\bar x_{c_j} + \theta_{c_j}, \sigma^2)
\end{align}
where $\theta_i$ is the effect for county $i$ (with prior mean $\mu$ and standard deviation $\tau$) and the $\beta$ are regression coefficients. The log Radon measurement in house $j$, $y_j$, depends on the effect $\theta_{c_j}$ for the county to which the house belongs, as well as features $z_{c_j}$ (the log uranium measurement in county $c_j$), $x_j$ (the floor of the house on which the measurement was taken), and $\bar x_{c_j}$ (the mean floor by county, a contextual effect). $\N^+(0, 1)$ indicates a Normal distribution with mean $0$ and variance $1$, truncated to nonnegative values.

\subsection{Baselines}
\subsubsection*{Mean Field ADVI}
The ADVI (MF) surrogate posterior is constructed with the same procedure as the ASVI posterior, but using only the $\vt{\alpha}$ parameters, or equivalently, fixing $\vt{\lambda} = 0$. As with ASVI, therefore, the surrogate posterior for each variable is in the same distribution family as its prior. This differs slightly from \citet{kucukelbir2017automatic}, in which surrogate posteriors are always bijectively transformed normal distributions, although we have no reason to believe that this difference is material to our experiments.

\subsubsection*{Inverse Autoregressive Flows}
Inverse Autoregressive Flows (IAFs) are normalizing flows which autoregressively transform a base distribution \citep{kingma2016improved} with a masked neural network \citep{papamakarios2017masked}.
We build an IAF posterior by transforming a standard Normal distribution with two sequential two-layer IAFs built with \protect\v{tfp.bijectors.MaskedAutoregressiveFlow}. The output of the flow is split and restructured to mirror the support of the prior distribution, and then constrained to the support of the prior (for example, by applying a sigmoid transformation to constrain values between zero and one, or a softplus to constrain values to be positive). In our experiments, we use
two different-sized IAF posteriors: the ``Large'' IAF has 512 hidden units in each layer and the ``Small'' IAF has 8 hidden units in each layer.

\subsubsection*{Multivariate Normal}
The MVN surrogate posterior is built by defining a full-covariance Multivariate Normal distribution with trainable mean and covariance, restructuring the support to the support of the prior, and constraining the samples to the prior support if necessary.

\subsubsection*{AR(1)}

The autoregressive model surrogate learns a linear Gaussian conditional between each pair of successive model variables:

\[x_{t+1} \sim \mathcal{N}(\vt{A}_t x_t + \vt{b}_t, \vt{D}_t)\]

where each $\vt{A}_t$ and $\vt{b}_t$ parameterize a learned linear transformation, and $\vt{D}_t$ is a learned diagonal variance matrix. The linear Gaussian autoregression operates on unconstrained values, which may then be then pushed through constraining transformations as required by the model. To stabilize the optimization, we omit direct dependence on global variables, i.e., when $\vt{x}_t$ is a global variable we fix $\vt{A}_t = \vt{0}$ (these are generally the first few variables sampled in each model).

\subsection{Training details}
\begin{figure}
\begin{subfigure}{0.33\textwidth}
\includegraphics[width=\textwidth]{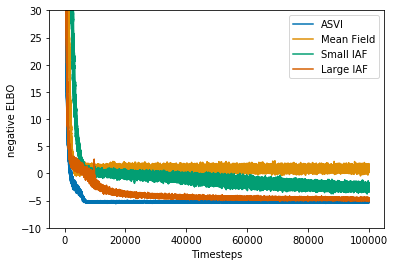}
\caption{Brownian motion (BR)}
\end{subfigure}
~
\begin{subfigure}{0.33\textwidth}
\includegraphics[width=\textwidth]{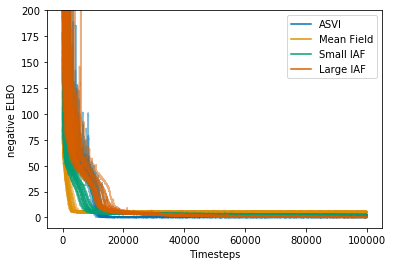}
\caption{Brownian motion with globals (BRG)}
\end{subfigure}
~
\begin{subfigure}{0.33\textwidth}
\includegraphics[width=\textwidth]{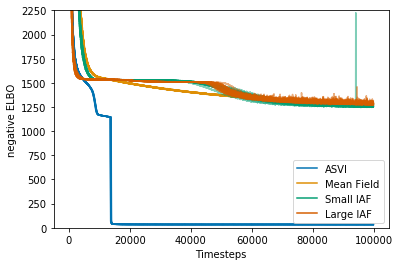}
\caption{Lorenz system (LZ) }
\end{subfigure}

\begin{subfigure}{0.33\textwidth}
\includegraphics[width=\textwidth]{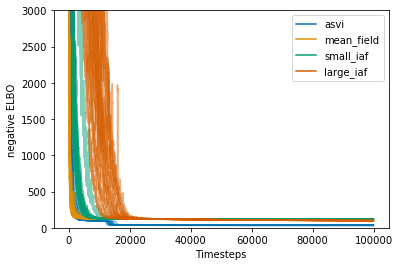}
\caption{Lorenz system with globals (LZG)}
\end{subfigure}
~
\begin{subfigure}{0.33\textwidth}
\includegraphics[width=\textwidth]{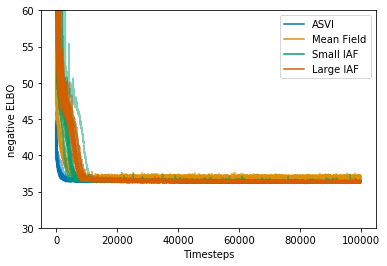}
\caption{Eight Schools (E)}
\end{subfigure}
~
\begin{subfigure}{0.33\textwidth}
\includegraphics[width=\textwidth]{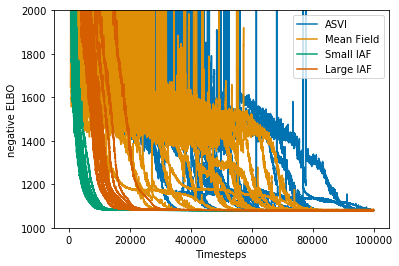}
\caption{Radon (R)}
\end{subfigure}
\caption{Training losses (negative ELBO values) for the ASVI, MF, Large IAF, and Small IAF baselines on the Inference Gym tasks. Each posterior was trained with the Adam
optimizer for 100000 steps.}
\label{fig:ig-losses}
\end{figure}

\begin{figure}
\begin{subfigure}{0.33\textwidth}
\includegraphics[width=\textwidth]{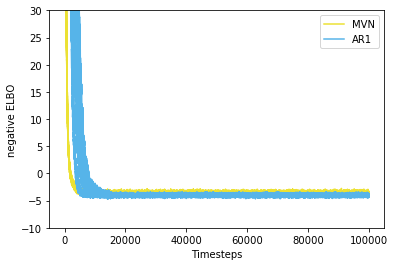}
\caption{Brownian motion (BR)}
\end{subfigure}
~
\begin{subfigure}{0.33\textwidth}
\includegraphics[width=\textwidth]{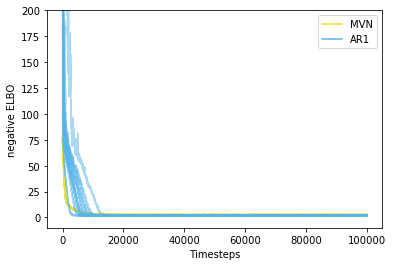}
\caption{Brownian motion with globals (BRG)}
\end{subfigure}
~
\begin{subfigure}{0.33\textwidth}
\includegraphics[width=\textwidth]{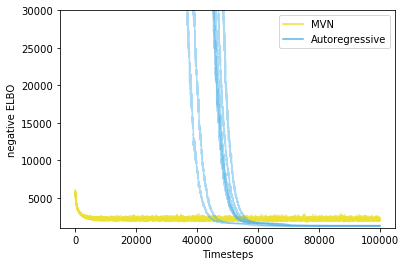}
\caption{Lorenz system (LZ) }
\end{subfigure}

\begin{subfigure}{0.33\textwidth}
\includegraphics[width=\textwidth]{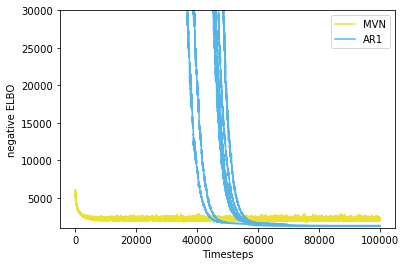}
\caption{Lorenz system with globals (LZG)}
\end{subfigure}
~
\begin{subfigure}{0.33\textwidth}
\includegraphics[width=\textwidth]{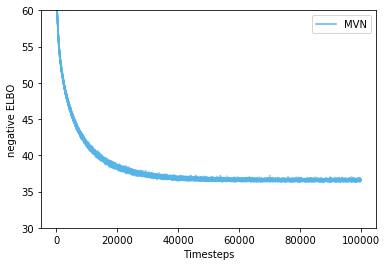}
\caption{Eight Schools (E)}
\end{subfigure}
~
\begin{subfigure}{0.33\textwidth}
\includegraphics[width=\textwidth]{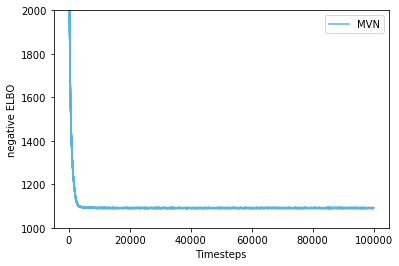}
\caption{Radon (R)}
\end{subfigure}
\caption{Training losses (negative ELBO values) for the MVN and AR(1) baselines on the Inference Gym tasks. Each posterior was trained with the Adam optimizer for 100000 steps.}
\label{fig:ig-losses-2}
\end{figure}

For each of the following inference tasks and posterior baselines, we fit a posterior using full-batch gradient descent
on a 1-sample Monte Carlo estimate of the ELBO. We use the Adam optimizer with a learning rates selected by hyperparameter sweep: 1e-2 learning rate for ASVI, MF, and AR(1), 1e-3 for the MVN and Small IAF, and 5e-5 for the Large IAF. Each posterior was trained for 100000 iterations; in \autoref{fig:ig-losses} and \autoref{fig:ig-losses-2}, we report the training curves for each posterior-task pair. We find that ASVI successfully converges in all tasks; in most cases, ASVI converges well before 100000 iterations, while MF, Large IAF, and Small IAF fail to converge to a good solution in a few of the tasks.

\section{Details of the neural SDE experiment}
\subsection{Models}
The function $\vt{F}(\vt{x})$ had the following form
\begin{equation}
    \vt{F}(\vt{x}) = W_2 \tanh{\left(\vt{x} + \tanh(W_1 \vt{x}) \right)}
\end{equation}
where $W_2$ and $W_1$ were $d \times d$ matrices whose entries were sampled in each of the $5$ repetitions from a centered normal with SD equal to $0.2$. Those matrices encodes the forward dynamical model and they were assumed to be known during the experiment. This is a Kalman filter-like setting where the form of the forward model is known and the inference is performed in the latent units. The neural SDE was integrated using Euler–Maruyama integration with step size equal to $1$ from $t = 0$ to $t = 9$. We trained the model by back-propagating though the integrator. 

We used two DCGAN generators as emission models. The networks were the DCGAN implemented in PyTorch. In the CIFAR experiment, we used the following architecture:
\begin{lstlisting}
ConvTranspose2d(100,64*8,4,1,0,
                bias=False),
BatchNorm2d(64*8),
ReLU(True)
ConvTranspose2d(64*8, 64*4,4,2,1,
                bias=False),
BatchNorm2d(ngf*4),
ReLU(True),
ConvTranspose2d(64*4,64*2,4,2,1,
                bias=False),
BatchNorm2d(64*2),
ReLU(True),
ConvTranspose2d(64*2,64,4,2,1,
                bias=False),
BatchNorm2d(ngf),
ReLU(True),
ConvTranspose2(64,4,kernel_size=1,
               stride=1,
               padding=0,
               bias=False),
Tanh()
\end{lstlisting}
The network pretrained on CIFAR was obtained from the GitHub repository: csinva/gan-pretrained-pytorch. The FashionGEN network was downloaded from the pytorch GAN zoo repository. The architectural details are given in \cite{radford2015unsupervised}.

\subsection{Baselines}
The ADVI (MF) baseline was obtained by replacing all the conditional Gaussian distributions in the probabilistic program with Gaussian distributions with uncoupled trainable mean and standard deviation parameters. ADVI (MN) was not computationally feasible in this larger scale experiment. Therefore, we implemented a a linear Gaussian model whith conditional densities:
\begin{equation}
    q(\vt{x}_t\mid \vt{x}_{t-1}) \sim \mathcal{N} \left(W\vt{x}_{t-1} + \vt{\alpha}_t, \vt{\sigma}^2_t  \right)~,
\end{equation}
where the matrix $W$, and the vectors $\vt{\alpha}_t$ and $\vt{\sigma}^2_t$ are learnable parameters.

\section{Details of the autoencoder experiment}
We used the following architectures in our deep autoencoder experiments.

Decoder 1 ($\vt{f}_1(\vt{z}_1)$)
\begin{lstlisting}
hidden_size=25
Linear(latent_size1, hidden_size)
ReLU()
Linear(hidden_size, latent_size2) 
\end{lstlisting}

Decoder 2 ($\vt{f}_2(\vt{z}_2)$)
\begin{lstlisting}
hidden_size = 75
Linear(latent_size2, hidden_size)
ReLU()
Linear(hidden_size, latent_size3) 
\end{lstlisting}

Decoder 3 ($\vt{g}_1(\vt{z}_3)$)
\begin{lstlisting}
Linear(latent_size3, image_size)
\end{lstlisting}

Decoder 4 ($\vt{\alpha}(\vt{y})$)
\begin{lstlisting}
Linear(latent_size3, image_size)
\end{lstlisting}

Inference network ($\vt{f}_1(\vt{z}_1)$)
\begin{lstlisting}
hidden_size1=120
Linear(image_size, hidden_size)
ReLU() # For hidden units
# Latent mean output 
# Latent log sd output
Linear(hidden_size1, latent_size3)  
Softplus() # For standard deviation output
hidden_size2=70
Linear(latent_size3, hidden_size2)
ReLU() # For hidden units
# Latent mean output 
Linear(hidden_size2, latent_size2)  
# Latent log sd output
Linear(hidden_size2, latent_size2)  
Softplus() # For standard deviation 
hidden_size3=70
Linear(latent_size2, hidden_size3)
ReLU() # For hidden units
# Latent mean output 
Linear(hidden_size3, latent_size1)  
# Latent log sd output
Linear(hidden_size3, latent_size1)  
Softplus() # For standard deviation 
\end{lstlisting}

\end{document}